\documentclass[10pt,twocolumn,letterpaper]{article}

\usepackage{wacv}
\usepackage{times}
\usepackage{epsfig}
\usepackage{graphicx}
\usepackage{amsmath}
\usepackage{amssymb}
\usepackage{color}
\usepackage{epsfig}
\usepackage{soul}
\usepackage{subfigure}
\usepackage{caption,booktabs,multirow,comment}

\usepackage[square,sort,comma,numbers]{natbib}
\usepackage[pagebackref=true,breaklinks=true,letterpaper=true,colorlinks,bookmarks=false]{hyperref}

\usepackage[font=small,skip=10pt]{caption}
\usepackage{array}
\newcolumntype{L}[1]{>{\raggedright\let\newline\\\arraybackslash\hspace{0pt}}m{#1}}
\newcolumntype{C}[1]{>{\centering\let\newline\\\arraybackslash\hspace{0pt}}m{#1}}
\newcolumntype{R}[1]{>{\raggedleft\let\newline\\\arraybackslash\hspace{0pt}}m{#1}}

\newcommand{\specialcell}[2][c]{%
  \begin{tabular}[#1]{@{}c@{}}#2\end{tabular}}

\wacvfinalcopy 

\def\wacvPaperID{376} 

\ifwacvfinal\fi
\begin{document}

\title{Towards Structured Analysis of Broadcast Badminton Videos}

\author{ Anurag Ghosh \hspace{2cm} Suriya Singh \hspace{2cm} C.V.Jawahar \\
CVIT, KCIS, IIIT Hyderabad \\
{\tt\small \{anurag.ghosh, suriya.singh\}@research.iiit.ac.in, jawahar@iiit.ac.in}
}

\maketitle
\ifwacvfinal\thispagestyle{empty}\fi

\begin{abstract}
Sports video data is recorded for nearly every major tournament but remains archived and inaccessible to large scale data mining and analytics. It can only be viewed sequentially or manually tagged with higher-level labels which is time consuming and prone to errors. In this work, we propose an end-to-end framework for automatic attributes tagging and analysis of sport videos. We use commonly available broadcast videos of matches and, unlike previous approaches, does not rely on special camera setups or additional sensors.

Our focus is on Badminton as the sport of interest. We propose a method to analyze a large corpus of badminton broadcast videos by segmenting the points played, tracking and recognizing the players in each point and annotating their respective badminton strokes. We evaluate the performance on 10 Olympic matches with 20 players and achieved 95.44\% point segmentation accuracy, 97.38\% player detection score (mAP@0.5), 97.98\% player identification accuracy, and stroke segmentation edit scores of 80.48\%. We further show that the automatically annotated videos alone could enable the gameplay analysis and inference by computing understandable metrics such as player's reaction time, speed, and footwork around the court, etc.
\end{abstract}

\section{Introduction}

Sports analytics has been a major interest of computer vision community for a long time. Applications of sport analytic system include video summarization, highlight generation~\cite{ghanem2012context}, aid in coaching~\cite{mlakar2017analyzing, remo2017technology}, player's fitness, weaknesses and strengths assessment, etc. Sports videos, intended for live viewing, are commonly available for consumption in the form of broadcast videos. Today, there are several thousand hours worth of broadcast videos available on the web. Sport broadcast videos are often long and captured in \textit{in the wild} setting from multiple viewpoints. Additionally, these videos are usually edited and overlayed with animations or graphics. Automatic understanding of broadcast videos is difficult due to its `unstructured' nature coupled with the fast changing appearance and complex human pose and motion. These challenges have limited the scope of various existing sports analytics methods.

\begin{figure}[t]
\begin{center}
\includegraphics[width=0.95\linewidth]{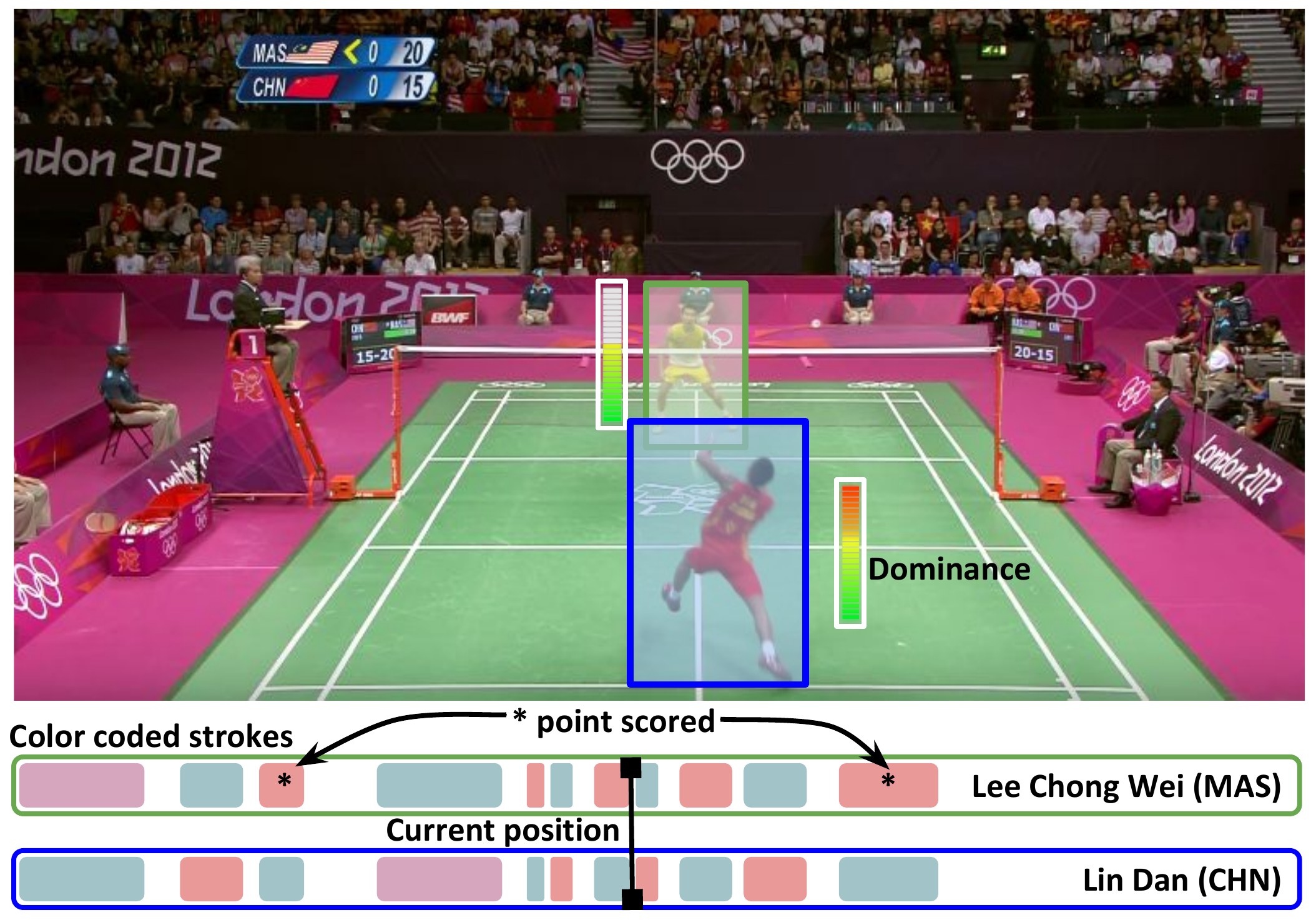}
\end{center}
\caption{We aim to automatically detect players, their tracks, points and strokes in broadcast videos of badminton games. This enables rich and informative analysis (reaction time, dominance, positioning, etc.) of each player at point as well as match level.}
\label{fig:motivation}
\end{figure}

Even today the analysis of sport videos is mostly done by human sports experts~\cite{dartfish} which is expensive and time consuming. Other techniques rely on special camera setup~\cite{remo2017technology} or additional sensors~\cite{mlakar2017analyzing} which adds to the cost as well as limits their utility. Deep learning based techniques have enabled a significant rise in the performance of various tasks such as object detection and recognition~\cite{simonyan2014very, ren2015faster, redmon2016you, he2016deep}, action recognition~\cite{wang2015action}, and temporal segmentation~\cite{lea2016temporal, singh2016first}.
Despite these advancements, recent attempts in sports analytics are not fully automatic for finer details~\cite{yan2014automatic, badminton2017chu} or have a human in the loop for high level understanding of the game~\cite{dartfish, badminton2017chu} and, therefore, have limited practical applications to large scale data analysis.


In this work, we aim to perform automatic annotation and provide informative analytics of sports broadcast videos, in particular, badminton games (refer to Fig.~\ref{fig:motivation}). We detect players, points, and strokes for each frame in a match to enable fast indexing and efficient retrieval. We, further, use these fine annotations to compute understandable metrics (e.g., player's reaction time, dominance, positioning and footwork around the court, etc.) for higher level analytics. Similar to many other sports, badminton has specific game grammar (turn-based strokes, winning points, etc.), well separated playing areas (courts), structured as a series of events (points, rallies, and winning points), and therefore, are suited well for performing analytics at a very large scale. There are several benefits of such systems. Quantitative scores summarizes player's performance while qualitative game analysis enriches viewing experience. Player's strategy, strengths, and weaknesses could be mined and easily highlighted for training. It automates several aspects of analysis traditionally done manually by experts and coaches.

Badminton poses different difficulties for its automatic analysis. The actions (or strokes) are intermittent, fast paced, have complex movements, and sometimes occluded by the other player. Further, the best players employ various subtle deception strategies to fool the human opponent. The task becomes even more difficult with unstructured broadcast videos. The cameras have an oblique or overhead view of players and certain crucial aspects such as wrist and leg movements of both players may not be visible in the same frame. Tracking players across different views makes the problem even more complicated. We discard the highlights and process only clips from \textit{behind the baseline} views which focus on both players and have minimal camera movements. For players detection, we rely on robust deep learning detection techniques. Our frame level stroke recognition module makes use of deep learned discriminative features within each player's spatio-temporal cuboid.

The major contributions of this paper are
\begin{enumerate}
\item We propose an end-to-end framework to automatically annotate badminton broadcast videos. Unlike previous approaches, our method does not rely on special camera setup or additional sensors.

\item Leveraging recent advancements in object detection, action recognition and temporal segmentaion, we predict game points and its outcome, players' tracks as well as their strokes.

\item We identify various understandable metrics, computed using our framework, for match and player analysis as well as qualitative understanding of badminton games.

\item  We introduce a large collection of badminton broadcast videos with match level point segments and outcomes as well as frame level players' tracks and their strokes. We use the official broadcast videos of matches played in London Olympics 2012.
\end{enumerate}
 
\pagebreak

\section{Related Work}

\begin{figure*}
\begin{center}
\includegraphics[width=0.9\linewidth]{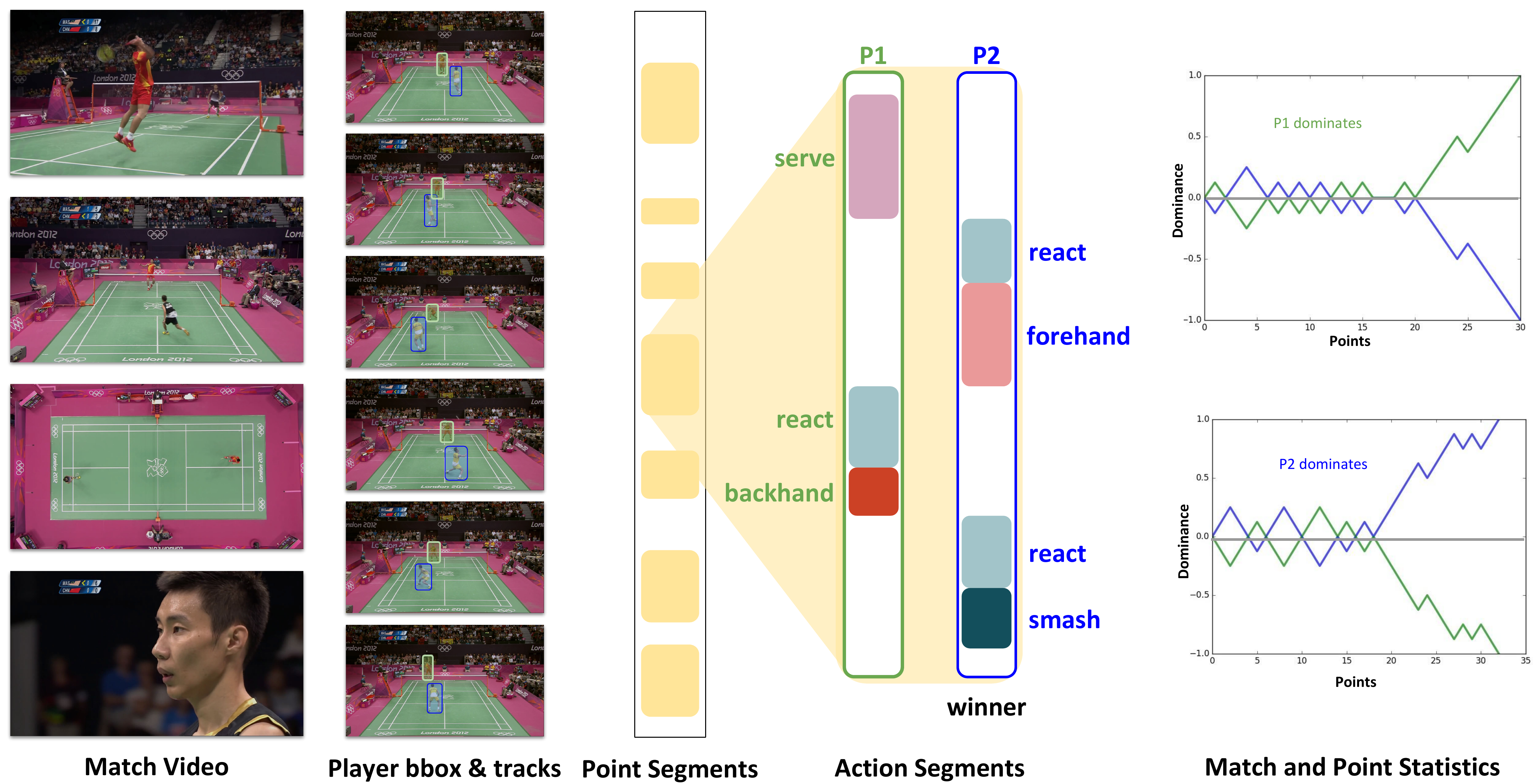}
\end{center}
   \caption{We propose to perform automatic annotation of various gameplay statistics in a badminton game to provide informative analytics. We model this task as players' detection and identification followed by temporal segmentation of each scored points. To enable deeper analytics, we perform dense temporal segmentation of player's strokes for each player independently.} 
\label{fig:overview}
\end{figure*}

\paragraph*{Sports Understanding and Applications:} Several researchers have worked on improving sports understanding using domain specific cues in the past~\cite{sha2014understanding, chen2014play}. Racket sports have received a lot of attention in this area with strides made in video summarization and highlight generation~\cite{hanjalic2003generic, ghanem2012context} and generating text descriptions~\cite{sukhwani2015tennisvid2text}. Reno et al.~\cite{remo2017technology} proposed a platform for tennis which extract 3D ball trajectories using a specialized camera setup. Yoshikawa et al.~\cite{yoshikawa2010automated} performed serve scene detection for badminton games with a specialized overhead camera setup. Zhu et al.~\cite{badminton2017chu} performed semi-automatic badminton video analysis by detecting the court and players, classifying strokes and clustering player strategy into offensive or defensive. Mlakar et al.~\cite{mlakar2017analyzing} performed shot classification while Bertasius et al.~\cite{amiaballer} assessed a basketball player's performance using videos from wearable devices. Unlike these approaches, our method does not rely on human inputs, special camera setup or additional sensors. Similar to our case, Sukhwani et al.~\cite{sukhwani2016frame} computed frame level annotations in broadcast tennis videos, however, they used a dictionary learning method to co-cluster available textual descriptions.

\paragraph*{Action Recognition and Segmentation:} 
Deep neural network based approaches such as Two Stream {\sc cnn}~\cite{simonyan2014two}, C3D~\cite{tran2014learning}, and it's derivatives~\cite{wang2015action, singh2016first} have been instrumental in elevating the benchmark results in action recognition and segmentation. {\sc rnn}s and {\sc lstm}s ~\cite{hochreiter1997long} have also been explored extensively ~\cite{lea2016segmental, fathi2013modeling} for this task owing to its representation power of long sequence data. Recently, Lea et al.~\cite{lea2016temporal} proposed temporal 1D convolution networks variants which are fast to train and perform competitively to other approaches on standard benchmarks for various temporal segmentation tasks.

In the context of sports activity recognition, Ramanathan et al.~\cite{ramanathan2016detecting} detected key actors and special events in basketball games by tracking players and classifying events using {\sc rnn}s with attention mechanism. Ibrahim et al.~\cite{ibrahim2016hierarchical} proposed to recognize multi-person actions in volleyball games by using {\sc lstm}s to understand the dynamics of players as well as to aggregate information from various players.

\paragraph*{Person Detection and Tracking:} An exhaustive survey of this area can be found in ~\cite{nguyen2016human}. Specific methods for sports videos~\cite{shitrit2011tracking, yan2014automatic, mentzelopoulos2013active} and especially for handling occlusions~\cite{held2016learning} have also been proposed in the past. In the context of applications involving player tracking data, Wang et al.~\cite{wang2016classifying} used tracking data of basketball matches to perform offensive playcall classification while Cervone et al.~\cite{cervone2014pointwise} performed point-wise predictions and discussed defensive metrics.


\pagebreak
\section{Badminton Olympic Dataset}

\begin{figure*}
\begin{center}
\includegraphics[width=0.8\linewidth]{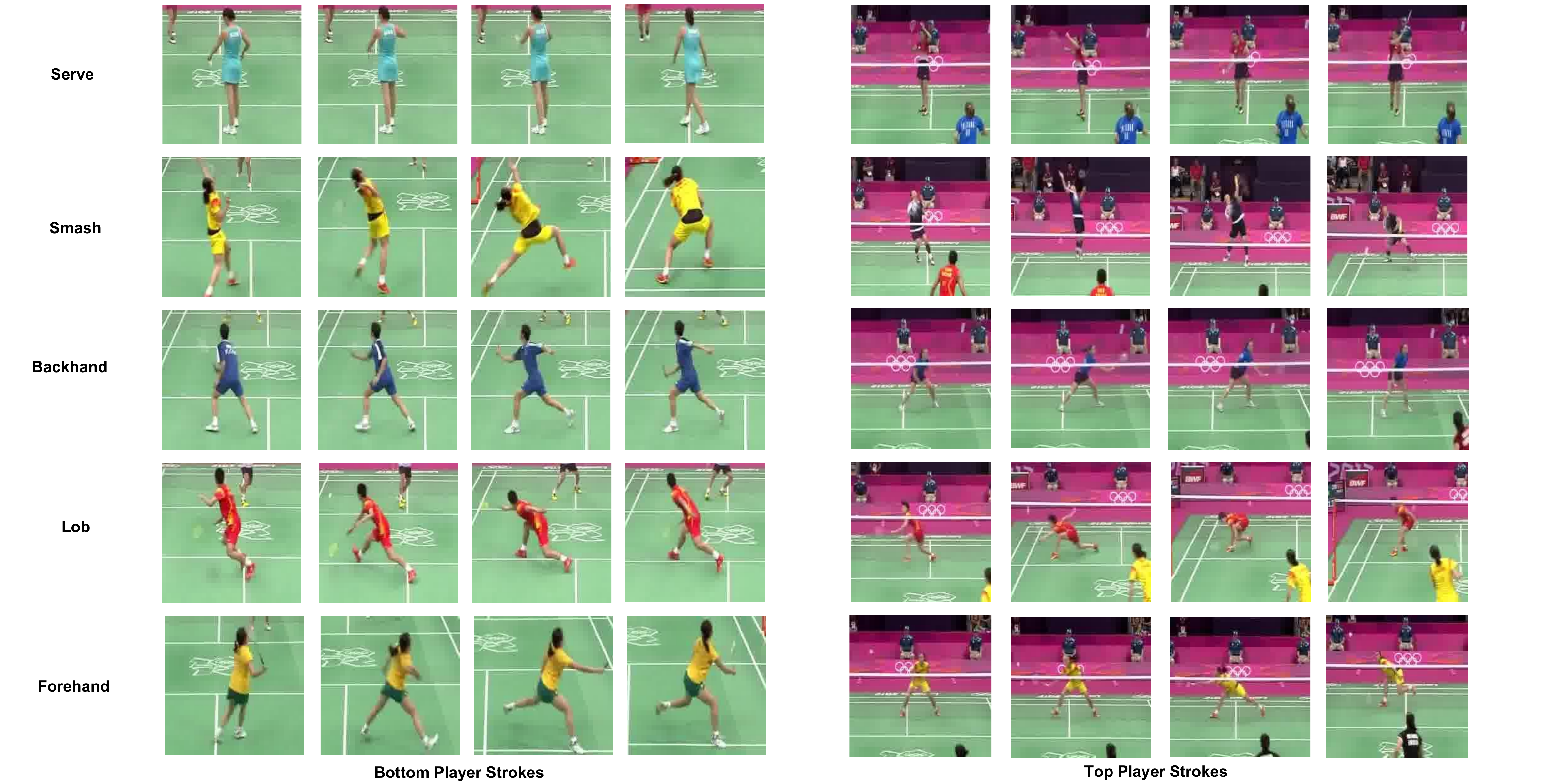}
\end{center}
   \caption{Representative ``strokes'' of bottom and top players for each class taken from our Badminton Olympic Dataset. The images have been automatically cropped using bounding boxes obtained from the player detection model. Top player appear smaller and have more complex background than the bottom player, therefore, are more difficult to detect and recognize strokes.}
\label{fig:qualresults}
\end{figure*}

\begin{table}[t]
    \centering
    \begin{tabular}{lcccc}
        \toprule[1.5pt]
        \specialcell{\bf Component} &\specialcell{\bf Classes} &\specialcell{\bf Total} &\specialcell{\bf Train} &\specialcell{\bf Test}  \\ \midrule
        Matches & {\sc na} & 10 & 7 & 3 \\
        Players & {\sc na} & 20 & 14 & 6 \\
        Player bboxes & 2 & 2988 & 2094 & 894 \\
        Point segments & 2 & 751 & 495 & 256 \\
        Strokes & 12 & 15327 & 9904 & 5423 \\
        \bottomrule[1.5pt]
    \end{tabular}
    \caption{Various statistics of our Badminton Olympic Dataset. Each match is typically one hour long. Train and test columns represents number of annotations used in respective split for experiments. Note that there is no overlap of players between train and test splits.}
    \label{table:dataset_stats}
\end{table}


We work on a collection of 27 badminton match videos taken from the official Olympic channel on YouTube\footnote{\url{https://www.youtube.com/user/olympic/}}. We focus on ``singles'' matches played between two players for two or three sets and are typically around an hour long. Statistics of the proposed dataset used in our experiments are provided in Table~\ref{table:dataset_stats}. Please refer to the supplementary materials for the full list of matches and the corresponding broadcast videos. We plan to release our dataset and annotations publicly post acceptance of the work.

\paragraph*{Matches: } To train and validate our approach, we manually annotate a subset of 10 matches. For this, we select only one match per player which means no player plays more than one match against any other player. We choose this criteria to incorporate maximum gameplay variations in our dataset as well as to avoid overfitting to any specific player for any of the tasks. We divide the 10 matches into training set of 7 matches and a test set of 3 matches. Note that this setup is identical to leave-{\sc n}-subjects-out criteria which is followed in various temporal segmentation tasks~\cite{fathi2011learning, stein2013combining, pirsiavash2012detecting, singh2016first}. Evaluation across pairs of unseen players also emphasize the generality of our approach.

\paragraph*{Points: } In order to localize the temporal locations of when points are scored in a match, we annotate 751 points and obtain sections that are corresponding to point and non-point segments. We annotate the current score, and the identity of the bottom player (to indicate the court switch after sets/between final set). Apart from this, we also annotate the serving and the winner of all the points in each set for validating outcome prediction.

\paragraph*{Player bounding boxes: } We focus on ``singles'' badminton matches of two players. The players switch court after each set and midway between the final set. In a common broadcast viewpoint one player plays in the court near to the camera while the other player in the distant court (see Fig.~\ref{fig:motivation}), which we refer to as bottom and top player respectively. We randomly sample and annotate 150 frames with bounding boxes for both players in each match (total around 3000 boxes) and use this for the player detection task. The players are occasionally mired by occlusion and the large playing area induces sudden fast player movements. As the game is very fast-paced, large pose and scale variations exist along with severe motion blur for both players.

\paragraph*{Strokes: } The badminton strokes can be broadly categorized as ``serve'', ``forehand'', ``backhand'', ``lob'', and ``smash'' (refer to Figure~\ref{fig:qualresults} for representative images). Apart from this we identify one more class, ``react'' for the purpose of player's gameplay analysis. A player can only perform one of five standard strokes when the shuttle is in his/her court while the opponent player waits and prepare for response stroke. After each stroke the time gap for response from other player is labeled as ``react''. Also, we differentiate between the stroke classes of the top player and the bottom player to identify two classes per stroke (say, ``smash-top'' and ``smash-bottom''). We also add a ``none'' class for segments when there is no specific action  occurring. We manually annotate all strokes of 10 matches for both players as one of the mentioned 12 ($5\cdot2 + 2$) classes.

The ``react'' class is an important and unique aspect of our dataset. When a player plays aggressively, that allows very short duration for the opponent to decide and react. It is considered to be advantageous for the player as the opponent often fails to react in time or make a mistake in this short critical time. To the best of our knowledge, ours is the only temporal segmentation dataset with such property due to the rules of the game. This aspect is evident in racket sports as a player plays only a single stroke (in a well separated playing space) at a time.

Sports videos have been an excellent benchmarks for action recognition techniques and many datasets have been proposed in the past~\cite{karpathy2014large, UCF101, ramanathan2016detecting, ibrahim2016hierarchical}. However, these datasets are either trimmed~\cite{karpathy2014large, UCF101} or focused on team based sports~\cite{ramanathan2016detecting, ibrahim2016hierarchical} (multi-person actions with high occlusions). On the contrary, for racket sports (multi-person actions with relatively less occlusion) there is no publicly available dataset. 
Our dataset is also significant for evaluation of temporal segmentation techniques since precise boundary of each action and processing each frame is of equal importance unlike existing temporal segmentation datasets~\cite{stein2013combining,fathi2011learning} which often have long non-informative background class. Sports videos also exhibit complex sequence of actions (depending on player's strategy, deceptions, injury etc.) unlike trivial sequences in other domains (e.g. cooking activity follows a fixed recipe) are excellent benchmarks for forecasting tasks.

\section{Extracting Player Data}

We start by finding the video segments that correspond to the play in badminton, discarding replays and other non-relevant sections. We then proceed to detect, track and identify players across these play segments. Lastly, we recognize the strokes played by the players in each play segment. We use these predictions to generate a set of statistics for effective analysis of game play.

\subsection{Point Segmentation}

\label{method:point_segmentation}
We segment out badminton ``points'' from the match by observing that usually the camera is behind the baseline during the play and involves no or minimal camera panning. The replays are usually recorded from a closer angle and focus more on the drama of the stroke rather than the game in itself (however, very few points are also recorded from this view), and thus adds little or no extra information for further analysis. We extract \textsc{hog} features from every $10^{th}$ frame of the video and learn a $\chi^2$ kernel {\sc svm} to label the frames either as a ``point frame'' or a ``non-point frame''. We use this learned classifier to label each frame of the dataset as a ``'point frame'' or otherwise and smoothen this sequence using a Kalman filter. 

\paragraph*{Evaluation} The \textsc{f1} score for the two classes for optimal parameters ($C$ and $order$) was found to be 95.44\%. The precision and recall for the point class were 97.83\% and 91.02\% respectively. 

\subsection{Player Tracking and Identification}

We finetune a FasterRCNN~\cite{ren2015faster} network for two classes, ``PlayerTop'' and ``PlayerBottom'' with manually annotated players bounding boxes. The ``top player'' corresponds to the player on the far side of the court and while the ``bottom player'' corresponds to the player on the near side of the court w.r.t to the viewpoint of the camera, and we'll be using this notation for brevity. For the rest of the frames, we obtain bounding boxes for both the players using the trained model. This approach absolves us from explicitly tracking the players with more complex multi-object trackers. We run the detector for every other frame in a point segment to get the player tracks.

We further find the players' correspondences across points, as the players change court sides after each set (and also in the middle of the third set). For performing player level analysis it is important to know which player's statistics we are actually computing. The players wear the same colored jersey across a match and it is dissimilar from the opponent's jersey. We segment the background from the foreground regions using moving average background subtraction method~\cite{heikkila2004real}. We then extract color histogram features from the detected bounding box after applying the foreground mask, and take it as our feature. Now, for each point, we randomly average 10 player features corresponding to the point segments to create 2 player features per point. We cluster the features using a Gaussian Mixture Model into 2 clusters for each match. We then label one cluster as the first player and the other cluster as the second player. This method, although simple, is not extensible to tournaments where both the players are wearing a standard tournament kit, such as Wimbledon in Tennis.

\paragraph*{Evaluation} For evaluating the efficacy of the learnt player detection model, we computed the $mAP@0.5$ values on the test set which were found to be 97.85\% for the bottom player, while it was found to be 96.90\% for the top player. 

For evaluating the player correspondences, we compare the identity assignments obtained from the clusters with the manual annotations of player identity for each point. As our method is unsupervised, we evaluated the assignments for all the points in 10 matches. The method described above yielded us an averaged accuracy of 97.98\%, which can be considered adequate for the further tasks.

\subsection{Player Stroke Segmentation}

\begin{table}[t]
    \centering
	\begin{tabular}{lrrrrr}
        \toprule[1.5pt]
        \specialcell{\bf ED-TCN} &\specialcell{\bf Metric} &\specialcell{\bf d=5} &\specialcell{\bf d=10} &\specialcell{\bf d=15} &\specialcell{\bf d=20}  \\ \midrule
        \multirow{2}{*}{\textsc{HOG}}  & Acc & 71.02 & 72.56 & 71.98 & 71.61 \\
             & Edit & 76.10 & 80.52 & 80.12 & 79.66 \\
        \hline
		\multirow{2}{*}{Spatial\textsc{CNN}}  & Acc & 69.19 & 68.92 & 71.31 & 71.49 \\
		     & Edit & 77.63 & 80.48 & 80.45 & 80.40 \\
        \toprule[1.5pt]
        \specialcell{\bf Dilated \textsc{TCN}} &\specialcell{\bf Metric} &\specialcell{\bf s=1} &\specialcell{\bf s=2} &\specialcell{\bf s=4} &\specialcell{\bf s=8}  \\ \midrule
        \multirow{2}{*}{\textsc{HOG}}  & Acc & 70.24 & 68.08 & 68.25 & 67.31 \\
             & Edit & 70.11 & 70.72 & 73.68 & 73.29 \\
        \hline
		\multirow{2}{*}{Spatial\textsc{CNN}}  & Acc & 69.59 & 69.75 & 69.37 & 67.03 \\
		     & Edit & 59.98 & 69.46 & 74.17 & 71.86 \\
        \bottomrule[1.5pt]
    \end{tabular}
    \caption{We evaluate the player stroke segmentation by experimenting with filter size `d' and sample rate 's' respectively. \textit{Acc} corresponds to per time step accuracy while \textit{Edit} corresponds to edit score.}
    \label{table:tcn_results}
\end{table}

\begin{figure}[t]
\begin{center}
\includegraphics[width=1\linewidth]{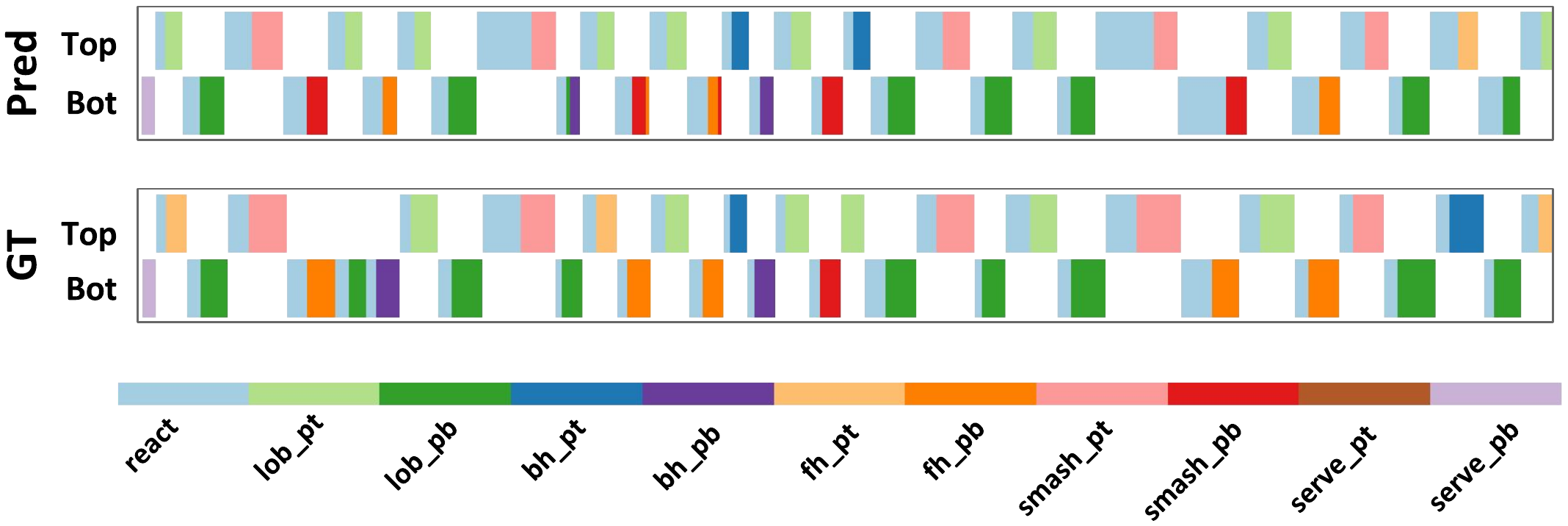}
\end{center}
\caption{\textbf{Stroke Visualization.} For a representative point, segment level strokes obtained from experiments are shown. Each action label has been color coded. \textit{(Best viewed in color)}}
\label{fig:errorvis}
\end{figure}

We employ and adapt the Temporal Convolutional Network (\textsc{tcn}) variants described by Lea et al.~\cite{lea2016temporal} for this task. The first kind, Encoder Decoder \textsc{tcn} (\textsc{ed-tcn}), is similar to the SegNet~\cite{badrinarayanan2015segnet} architecture (used for semantic segmentation tasks), the encoder layers consist of, in order, temporal constitutional filters, non linear activation function and temporal max-pooling.  The decoder is analogous to the encoder instead it employs up-sampling rather than pooling, and the order of operations is reversed. The filter count of each encoder-decoder layer is maintained to achieve symmetry w.r.t. architecture. The prediction is then the probability of each class per time step obtained by applying the softmax function.

The second kind, Dilated \textsc{tcn} is analogous to the WaveNet~\cite{oord2016wavenet} architecture (used in speech synthesis tasks). A series of blocks are defined (say $B$), each containing $L$ convolutional layers, with the same number of filters $F_{w}$. Each layer has a set of dilated convolutions with rate parameter $s$, activation and residual connection that combines the input and the convolution signal, with the activation in the $l^{th}$ layer and the $j^{th}$ block is denoted as $S^{(j,l)}$. Assuming that the filters are parameterized by $W^{(1)}$, $W^{(2)}$, $b$ along with residual weight and bias parameters $V$ and $e$,

\begin{align*}
\hat{S}_{t}^{(j,l)} &= f(W^{(1)}S_{t-s}^{j, l-1} + W^{(2)}S_{t}^{j, l -1} + b) \\
S_{t}^{(j,l)} &= S_{t}^{(j,l-1)} + V\hat{S}_t^{(j,l)} + e
\end{align*}

The output of each block is summed using a set of skipped connections by adding up the activations and applying the $ReLU$ activation, say $Z^{0}_{t} = ReLU(\sum_{j=1}^{B} S_(t)^{(j,L)}$. A latent state is defined as $Z^{(1)}_{t} = ReLU(V_{r}Z_{t}^{(0)} + e_{r})$ where $V_{r}$ and $e_{r}$ are learned weight and bias parameters. The predictions are then given by applying the softmax function on $Z_{t}^{(1)}$.  

To learn the parameters, the loss employed is categorical cross-entropy with {\sc sgd} updates. We use balanced class weighting for both the models in the cross entropy loss to reduce the effect of class imbalance.

\begin{figure}[t]
\begin{center}
\includegraphics[width=1.0\linewidth]{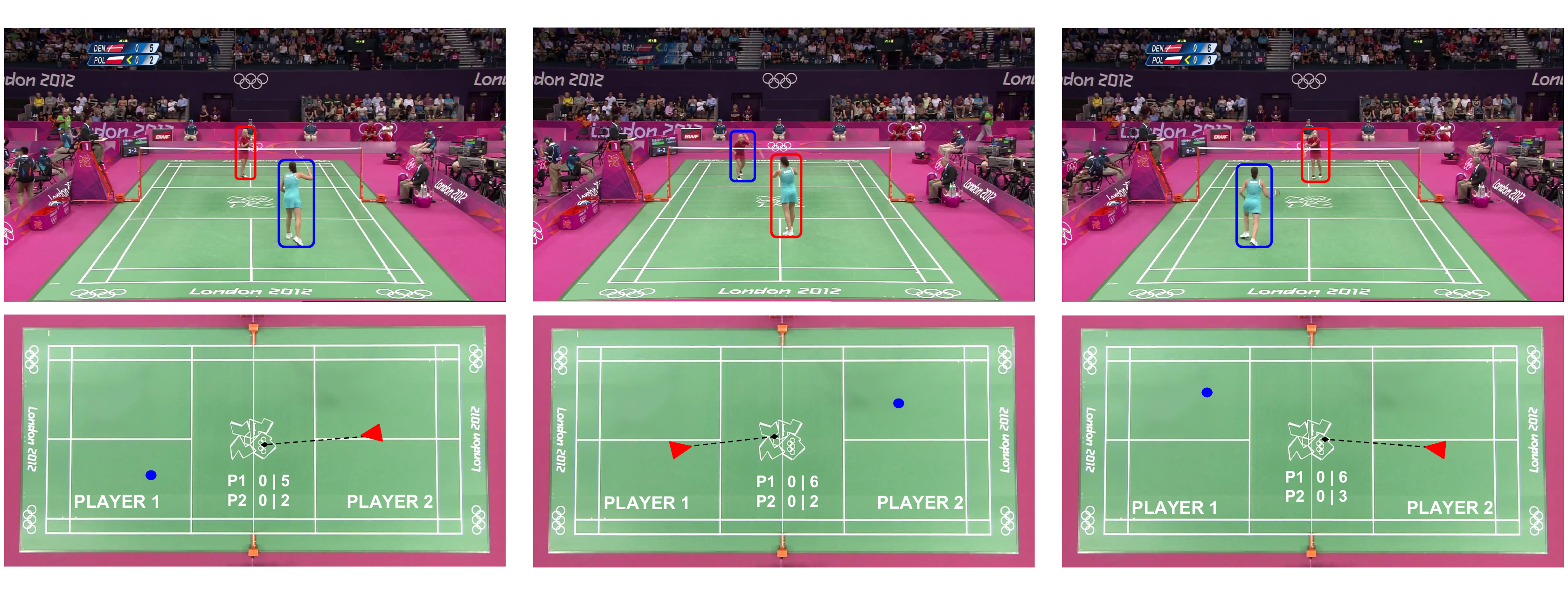}
\end{center}
\caption{\textbf{Detecting Point Outcome.} The serving player is indicated in red. It can be observed that the serving player is usually closer to the mid-line than the receiver who centers himself in the opposite court. Also, the positions of the players in the second point w.r.t. to the first point indicate that the bottom player has won the last point (as the serve is switched). \textit{(Best viewed in color)}}
\label{fig:badmintonscore}
\end{figure}

We experiment with two different feature types, \textsc{hog} and a \textsc{cnn}. Inspired by the use of \textsc{hog} features by~\cite{badminton2017chu} for performing stroke recognition, we extract \textsc{hog} features. As~\cite{lea2016temporal} benchmark using trained Spatial \textsc{cnn} features and the current benchmarks all use convolutional neural networks, we employ the Spatial \textsc{cnn} from the two stream convolution network model~\cite{simonyan2014two} to extract features. However, instead of extracting features globally, we instead utilize the earlier obtained player tracks and extract the image region of input scale ($454\times340$) centered at player track centroid for each time step which are then resized to $224\times224$. We then independently extract features for both the players and concatenate the obtained features per frame. The Spatial \textsc{cnn} used is trained on the \textsc{ucf101} dataset, and we experiment with the output of \textsc{fc7} layer as our features. The \textsc{hog} features are extracted with a cell size of 64.

We use the default parameters for training the \textsc{tcn} variants as reported in~\cite{lea2016temporal}. We experiment the effect of dilation by setting the sample rate ($s$) at 1,2,4 and 8 fps for Dilated \textsc{tcn}. For \textsc{ed-tcn} we vary the convolutional filter size (say $d$) of 5, 10, 15 and 20 (setting $s$ = 2). We employ acausal convolution for \textsc{ed-tcn} by convolving from $X_{t-\frac{d}{2}}$ to $X_{t+\frac{d}{2}}$. For the Dilated \textsc{tcn} case, we add the term $W^{(2)}S_{t + s}^{j, l -1}$ to the update equation mentioned earlier~\cite{lea2016temporal}. Here, $X$ is the set of features per point and $t$ is the time step.

\paragraph*{Evaluation} We employ the per frame accuracy and edit score metric used commonly for segmental tasks~\cite{lea2016temporal}, and the results can be seen in Table.~\ref{table:tcn_results}. We also experimented with causal models by convolving from $X_{t-d}$ to $X_{t}$ but observed that the performance of those models is not comparable to acausal models and thus did not report those results. The low performance of causal models can be attributed to the fact that badminton is fast-paced and unpredictable in nature. \textsc{ed-tcn} outperforms Dilated \textsc{tcn} which is consistent with benchmarking on other datasets~\cite{lea2016temporal}. We can observe that the filter size of $10$ is most appropriate for the \textsc{ed-tcn} while the sample rate of $4$ is most appropriate for Dilated \textsc{tcn}. From Fig~\ref{fig:errorvis}, it can be seen that the backhand and forehand strokes are prone to confusion, also smash and forehand strokes. Please refer to the supplementary materials for more exhaustive and detailed results. 

\section{Detecting Point Outcome}

\begin{figure}[t]
\begin{center}
\includegraphics[width=1.0\linewidth]{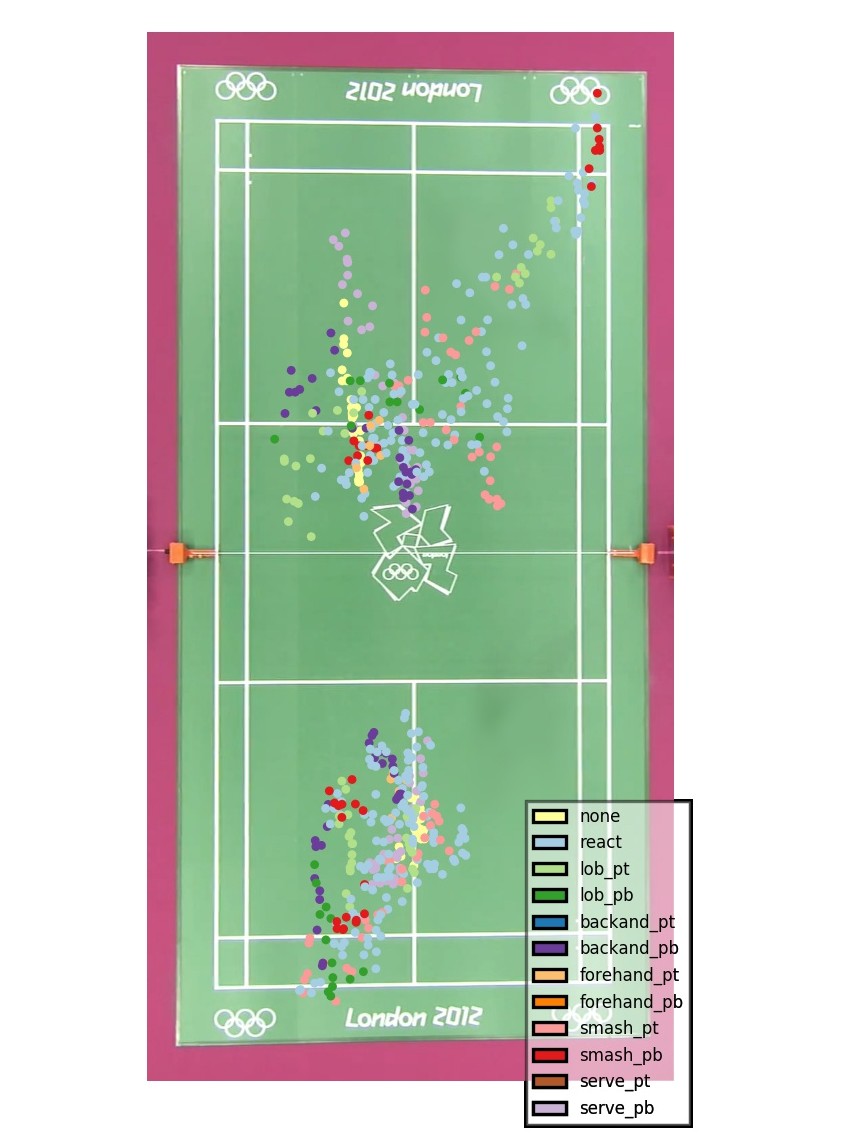}
\end{center}
\caption{\textbf{Point Summary.}  
We show the frame level players' positions and footwork around the court corresponding to game play of a single point won by the bottom player. The color index correspond to the stroke being played. Note that, footwork of bottom player is more dense compared to that of top player indicating the dominance of bottom player.
\textit{(Best viewed in color)}}
\label{fig:pointsummary}
\end{figure}

\begin{figure*}
\begin{center}
\includegraphics[width=0.9\linewidth]{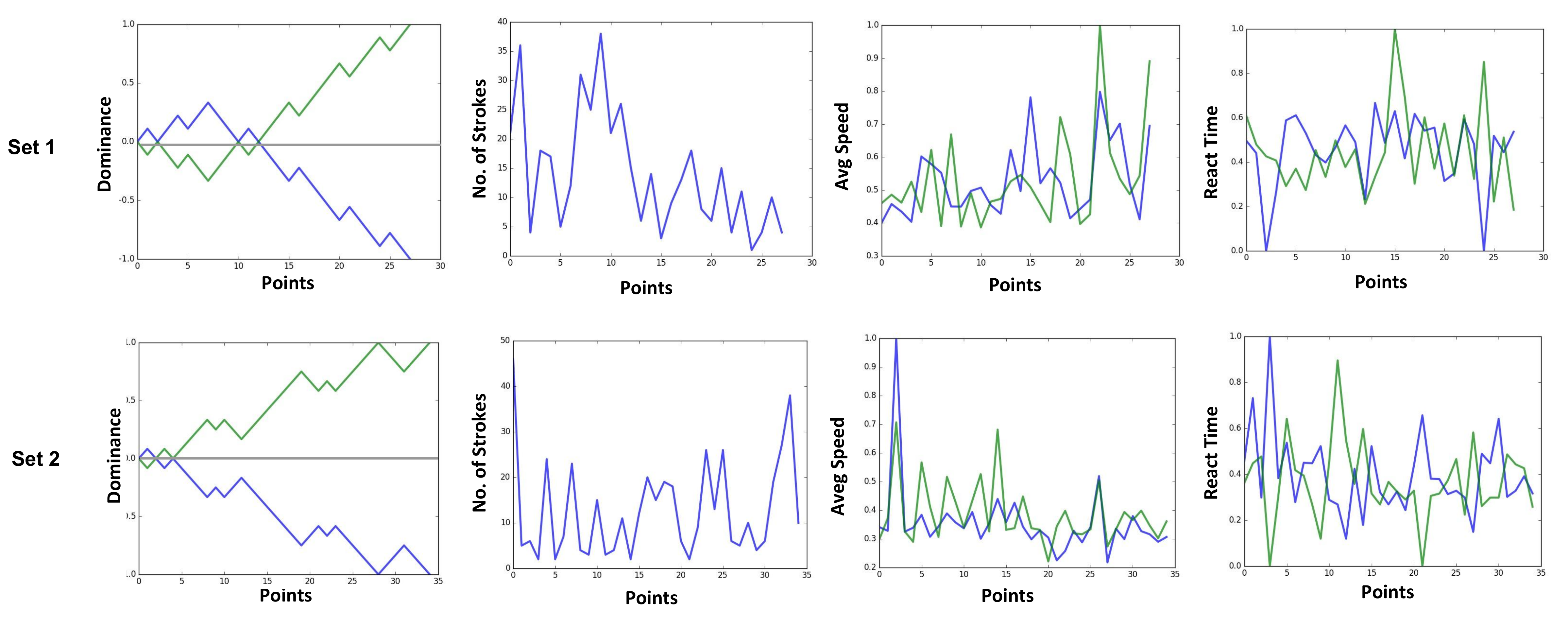}
\end{center}
   \caption{The computed statistics for a match, where each row corresponds to a set. It should be noted that green corresponds to the first player, while blue corresponds to second player. The first player won the match. \textit{(Best viewed in color)}}
\label{fig:badminstats}
\end{figure*}

The badminton scoring system is simple to follow and incorporate into our system. At the beginning of the game (score 0 --- 0) or when the serving player's score is even, the serving player serves from the right service court, otherwise, from the left service court. If the serving player wins a rally, they score a point and continues to serve from the alternate service court. If the receiving player wins a rally, the receiving player scores a point and becomes the next serving player.

We exploit this game rule and its relationship to players' spatial positions on the court for automatically predicting point outcomes.
Consider point video segments obtained from point segmentation (Section ~\ref{method:point_segmentation}), we record predicted players' positions, and strokes played in each frame. At the start of next rally segment, the player performing the ``serve'' stroke is inferred as the winner of previous rally and point is awarded accordingly. We, therefore, know the point assignment history of each point by following this procedure from the start and until the end of the set. Also, it should be noted that for a badminton game the spatial position of both the serving and the receiving players is intrinsically linked to their positions on the court (See Fig.~\ref{fig:badmintonscore} for a detailed explanation). While similar observations have been used earlier by ~\cite{yoshikawa2010automated} to detect serve scenes, we detect both the serving player and the winning player (by exploiting the game rules) without any specialized setup.

We formulate the outcome detection problem as a binary classification task by classifying who is the serving player, i.e. either the top player is serving or the bottom player is serving. Thus, we employ a kernel \textsc{svm} as our classifier, experimenting with polynomial kernel. The input features are simply the concatenated player tracks extracted earlier i.e. for the first $k$ frames in a point segment, we extract the player tracks for both the players to construct a vector of length $8k$. 

We varied the number of frames and the degree of polynomial kernel for our experiments and tested on the 3 test matches as described earlier. We observed that the averaged accuracy was found to be $94.14\%$ when the player bounding boxes of the first $50$ frames are taken and the degree of the kernel is six.

\section{Analyzing Points}

The player tracks and stroke segments can be utilized in various ways for data analysis. For instance, the simplest method would be the creation of a pictorial \textbf{point summary}. For a given point (see Fig.~\ref{fig:pointsummary}), we plot the ``center-bottom'' bounding box positions of the players in the top court coordinates by computing the homography. We then color code the position markers depending on the ``action''/``reaction'' they were performing then. From Fig.~\ref{fig:pointsummary}, it is evident that the bottom player definitely had an upper hand in this point as the top player's positions are scattered around the court. These kind of visualizations are useful to quickly review a match and gain insight into player tactics.

We attempt to extract some meaningful statistics from our data. The temporal structure of Badminton as a sport is characterized by short intermittent actions and high intensity~\cite{phomsoupha2015science}. The pace of badminton is swift and the court situation is always continuously evolving, and difficulty of the game is bolstered by the complexity and precision of player movements. The decisive factor for the games is found to be speed~\cite{phomsoupha2015science}, and it's constituents,
\begin{itemize}
\setlength\itemsep{0.2em}
\item[--] Speed of an individual movement
\item[--] Frequency of movements
\item[--] Reaction Time
\end{itemize}

In light of such analysis of the badminton game, we define and automatically compute relevant measures that can be extracted to quantitatively and qualitatively analyze player performance in a point and characterize match segments. 
We use the statistics presented in Fig.~\ref{fig:badminstats} for a match as an example.

\begin{enumerate}
\item \textbf{Set Dominance} We utilize the detected outcomes and the player identification details to define dominance of a player. We start the set with no player dominating over the other and we define a player as dominating if they have won consecutive points in a set and add one mark to the dominator and subtract one mark from the opponent likewise. We then plot the time sequence to find both ``close'' and ``dominating'' sections of a match.

For instance, in Fig.~\ref{fig:badminstats}, which are statistics computed for a match, it's apparent that the initial half of the first set was not dominated by either players and afterwards one of the players took the lead. The same player continued to dominate in the second set and win the game.

\item \textbf{Number of strokes in a point} A good proxy for aggressive play is the number of strokes being played by the two players. Aggressive and interesting play usually results in long rallies and multiple back-and-forth plays before culminating in a point scored for one or the other players. To approximate, we count the number of strokes in a point. Interestingly, it can be observed in Fig.~\ref{fig:badminstats} that the stroke count is higher during the points none of the players are dominating in the match we have taken as example.

\item \textbf{Average speed in a point} To find the average speed of the players in a point, we utilize our player tracks. However, displacement of both the players would manifest differently in the camera coordinates. Thus we detect the court lines in the video frame and find the homography with the camera view (i.e. behind the baseline view) of the court. We then use the bottom of the player bounding boxes as proxy for feet and track that point in the camera view. Using a Kalman Filter, we compute displacement and thus speed in the overhead view (taking velocity into account through the observations matrix) and normalize the values. This would act as a proxy for intensity within a point.

\item \textbf{Average Reaction Time} We approximate reaction time by averaging the time for react class separately for both the players and then normalizing the values. We assume that the reaction time for the next stroke corresponds to the player who is performing it (See Fig.~\ref{fig:errorvis}) to disambiguate between the reactions. This measure could be seen as the leeway the opponent provides the player.

\end{enumerate}

\section{Discussions}
In this work, we present an end-to-end framework for automatic analysis of broadcast badminton videos. We build our pipeline on off-the-shelf object detection~\cite{ren2015faster}, action recognition and segmentation~\cite{lea2016temporal} modules. Analytics for different sports rely on these modules making our pipeline generic for various sports, especially racket sports (tennis, badminton, table tennis, etc.). Although these modules are trained, fine-tuned, and used independently, we could compute various useful as well as easily understandable metrics, from each of these modules, for higher-level analytics. The metrics could be computed or used differently for different sports but the underlying modules rarely change. This is because broadcast videos of different sports share the similar challenges. We present a comprehensive analysis of errors and failures of our approach in the supplementary materials. We also show a detailed match level stroke, players' positions and footwork visualizations as well as match statistics obtained using our framework along with videos to showcase the information augmentation during game play.



Rare short term strategies (e.g., deception) and long term strategies (e.g., footwork around the court) can be inferred with varying degree of confidence but not automatically detected in our current approach. To detect deception strategy, which could fool humans (players as well as annotator), a robust fine-grained action recognition technique would be needed. Whereas, predicting footwork requires long term memory of game states. These aspects of analysis are out of scope of this work. 
Another challenging task is forecasting player's reaction or position. It's specially challenging for sports videos due to the fast paced nature of game play, complex strategies as well as unique playing styles.

\pagebreak

{\small
\bibliographystyle{ieee}
\bibliography{egbib}
}

\end{document}